\documentclass[10pt,twocolumn,letterpaper]{article}

\usepackage{wacv}
\usepackage{times}
\usepackage{epsfig}
\usepackage{graphicx}
\usepackage{amsmath}
\usepackage{amssymb}

\usepackage{booktabs} 
\usepackage{color}
\usepackage{enumitem}
\usepackage{hyperref}

\newcommand{\jose}[1]{\textcolor{black}{#1}}

\wacvfinalcopy 

\def\wacvPaperID{458} 

\begin{document}

\title{An Analysis of Human-Centered Geolocation}

\author{
Kaili Wang\footnotemark[1] \hspace{1cm} 
Yu-Hui Huang\footnotemark[1] \hspace{1cm}
Jose Oramas M.\footnotemark[1] \hspace{1cm}
Luc Van Gool\footnotemark[1] \footnotemark[2]  \hspace{1cm}
Tinne Tuytelaars\footnotemark[1]
\\
{~}\footnotemark[1]{~}\textit{KU Leuven, ESAT-PSI, imec}   \hspace{1cm} \footnotemark[2]{~}\textit{ETH/D-ITET/CVL} \\
}

\maketitle
\ifwacvfinal\thispagestyle{empty}\fi

\begin{abstract}
Online social networks contain a constantly increasing amount 
of images - most of them focusing on people. 
Due to cultural and climate factors, fashion trends and physical 
appearance of individuals differ from city to city. 
In this paper we investigate to what extent such cues 
can be exploited in order to infer the geographic location, 
i.e. the city, where a picture was taken.
We conduct a user study, as well as 
an evaluation of automatic methods based on convolutional neural 
networks.
Experiments on the Fashion 144k and a Pinterest-based dataset show
that the automatic methods succeed at this task  to a reasonable extent. As a matter of fact, our empirical results suggest that 
automatic methods can surpass human performance by a large margin.
Further inspection of the trained models shows that human-centered characteristics, like clothing style, physical 
features, and accessories, are informative for the task at hand.
Moreover, it reveals that
also contextual features, e.g. wall type, natural environment, etc., 
are taken into account by the automatic methods.
\end{abstract}



\section{Introduction}

The increasing amount of low-cost camera-capable devices released on 
the market and the popularity of online social networks have produced 
an almost exponential increase in the amount of visual data uploaded 
to the Web. 
A large subset of this data consists of ``human-centered images'', i.e. images 
whose content is mostly focused on a single individual. A side effect 
of this human-centered characteristic is that the amount of background information is 
reduced, thus, limiting the possibilities of inferring the location 
where the image was taken in a direct manner, i.e. by recognizing the place.
Two questions then arise: Is it still possible to geolocate the image?, 
and if so, what are the useful visual cues for this task?.

Here we start from the hypothesis that cultural and climate factors 
have an influence on the fashion trends and physical 
appearance of individuals of different countries.
For example, individuals from tropical locations are more likely 
to have a tanned skin color than those living in polar regions. As to their clothing, people near the poles 
are more likely to wear warm clothing than individuals living near 
the Equator. Likewise, people at seaside towns may dress differently 
than those in dense urban areas.
%
We investigate to what extent such cues
can be exploited to predict the geographic location, 
i.e. the city, where a picture was taken.

\begin{figure}
\centering
\includegraphics[width=0.5\textwidth]{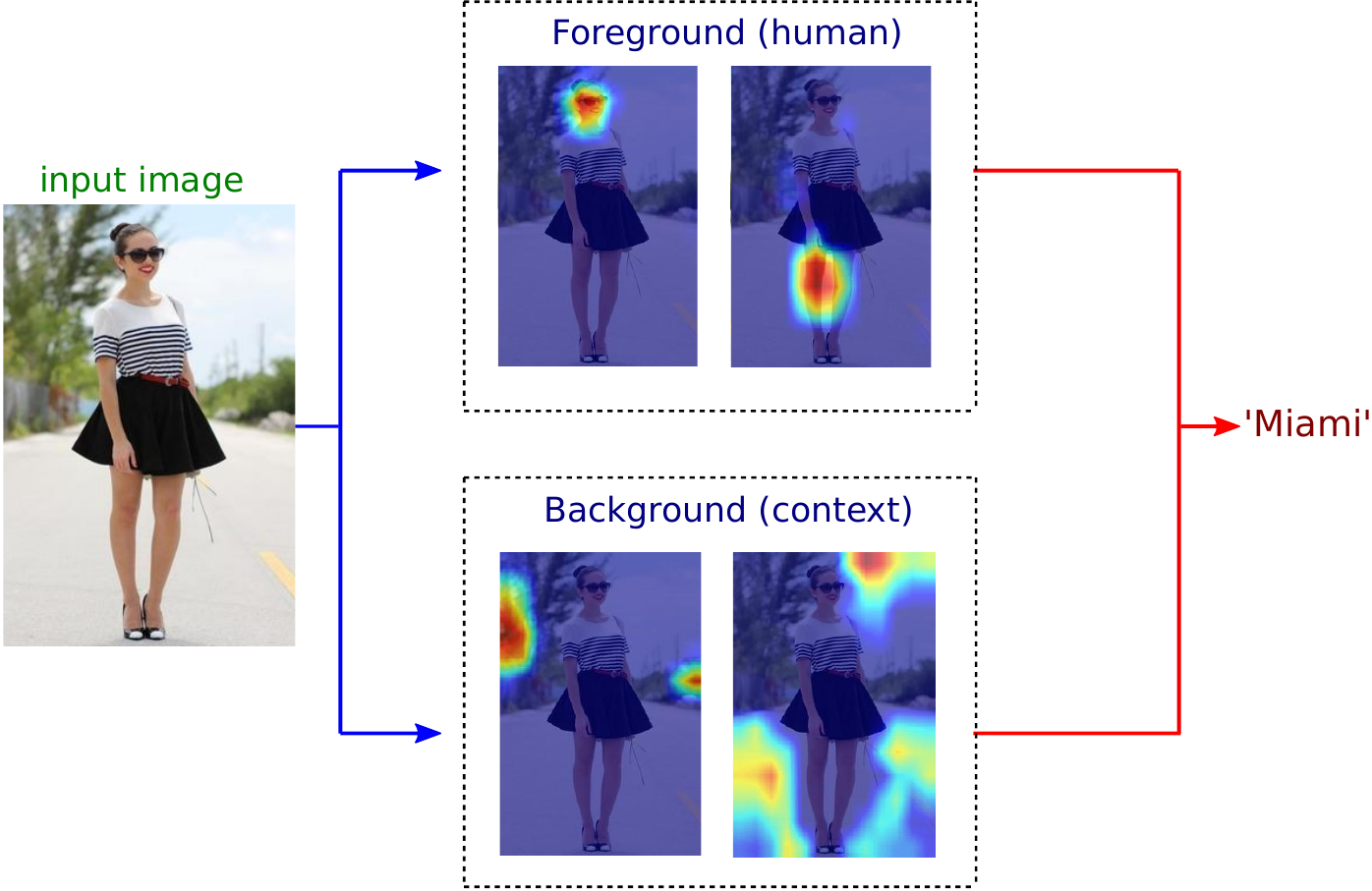}

\caption{
Can we infer the location (city) where a human-centered picture, such as the one shown on the left, was taken ? And where do the visual features used in this process come from? 
}
\label{fig:geolocationTeaser}
\end{figure}

We formulate the geolocation problem as a classification 
task where the city names are the classes to be inferred 
from human-centered images.
Firstly, we let people guess, serving as a baseline, providing 
an idea about human performance, and illustrating the difficulty 
of the task. 
Secondly, we conduct a series of experiments with automatic 
classification methods based on Convolutional Neural Networks (ConvNet). 
Finally, we analyze the origin of the visual features learned by the ConvNet-based 
methods (Fig.~\ref{fig:geolocationTeaser}). More specifically, we verify 
whether the learned features are focused on the foreground (human-region) 
or the background (context).
Our automatic methods surpass human performance by a large margin for this task.
Moreover, our analysis suggests that human-centered 
features are being used.  
In addition, despite their less dominant nature, 
contextual features, e.g. wall type, natural 
environment, etc., are taken into account by the automatic methods.

Being able to geolocate human-centered images is not just of academic interest. 
Knowing the relation between geographic location and clothing is of commercial 
importance as well. For example, online shops can leverage this 
type of information to provide geography-based recommendations. 
Likewise, multinational retailers can use it 
to decide which type of products to put on their shelves.

\vspace{2mm}
\noindent The main contributions of this paper are two-fold:
\vspace{-2mm}
\begin{itemize}[leftmargin=*]
\item Our extensive experiments provide significant empirical evidence of the 
feasibility of geolocalization from human-centered images. 
To the best of our knowledge this is the first work addressing the visual geolocalization 
problem from a human-centered perspective.
\vspace{-2mm}
\item In addition to reporting quantitative performance of several automatic methods, 
we propose an inspection method in order to identify the origin of the features 
learned by the network.
%
\end{itemize}

This paper is organized as follows: in Section~\ref{sec:relatedWork} 
we position our work w.r.t. earlier work. Section~\ref{sec:analysis} 
presents the methodology followed in our analysis.
In Sections~\ref{sec:evaluation} \& \ref{sec:discussion}, we 
conduct a series of experiments and discuss the observations and 
findings made throughout our evaluation. 
Finally, in Section~\ref{sec:conclusion}, we draw conclusions from 
the analysis.


\section{Related Work}
\label{sec:relatedWork}
We position our work w.r.t. related work on the topics of photo geolocation, fashion analysis, and inspection of ConvNets. 

\textbf{Photo geolocation} Photo-based geolocation has been studied from different perspectives. Some authors focus on landmarks of cities \cite{Avrithis10, Quack08,Zheng09}, some on street view images \cite{Chen11,Kim16,Amir14}, and some on arbitrary photos \cite{Hays08,Hays15,Weyand16}. These contributions can be divided into two groups: a first coming from retrieval \cite{Avrithis10,Chen11,Hays08,Hays15,VodeepIM2GPS}, and a second from classification~\cite{Weyand16}. We treat our problem as a classification task and employ a ConvNet instead of handcrafted features. 
Differently from \cite{Weyand16}, we predict city-related classes instead of discrete GPS coordinates. Moreover, we target human-centered photos.

 \textbf{Fashion analysis}
 Simo-Serra et al. \cite{SimoSerra15} exploit a large amount of data from \textit{chictopia.com}, a large social network for fashion style sharing. Based on user posts and tags, they train a model to predict how fashionable a person looks from a photo and suggest a way to dress better. 
Similarly, Bossard et al. \cite{BossardDLWQG12} train a random forest to classify the clothing style of people in natural images.
Murillo et al. \cite{Murillo12} and Wang et al. \cite{Wang11} predict a person's occupation 
based on the clothing cues and contexts from a photo. Wang et al. apply hand-crafted features to represent human body parts. Using sparse coding \cite{Liu09,Yuan12}, they learn representative patterns for each occupation. Similar to us, they consider foreground and background information for the classification. 


\textbf{Inspecting ConvNets}
ConvNets have been shown to be powerful tools for feature representation. However, without further analysis they remain somewhat of a black box. Simonyan et al. \cite{Simonyan13} tackled this problem by finding the images that activate specific ConvNet nodes. Similarly, Zeiler et al. \cite{Zeiler14} proposed a DeconvNet-based network to visualize activations. Later, Springenberg
et al. \cite{Springenberg14} introduced a new variant of DeconvNet-based approach called guided backpropagation for feature visualization. Compared with the DeconvNet, the guided backpropagation provides a sharper and cleaner visualization for higher layers of the network.
\cite{zhou2015cnnlocalization} proposes a weighted sum over the spatial locations of the activations of the filters from the last convolutional layer to generate an activation map. Finally, a heatmap is generated by upsampling the activation map to the size of the input image. Similarly, \cite{netdissect2017} proposed to exhaustively 
match the internal upsampled activations of every filter from the convolutional layers
against a dataset with pixel-wise annotated concepts in order to measure interpretability.
Similar to \cite{netdissect2017, zhou2015cnnlocalization}, we use upsampled activations from the last convolutional layer as means to identify the features that the network has learned. Moreover, we exploit the spatial information encoded in this layer in order to verify whether the features considered by the network come from the persons in the images or from the background.

%


\section{Methodology}
\label{sec:analysis}

As said earlier, we formulate the geolocation problem as a classification problem, with the goal of assigning to a given image $I$ a city class label, from a fixed set of cities $C$.
Following the landmark work of \cite{Krizhevsky12}, we address this 
problem through Convolutional Neural Networks (ConvNets).
In the following sections we describe several ConvNet-based image classification schemes.

\subsection{Image Classification with ConvNets}
\label{sec:imageClassification}


A deep ConvNet is a powerful mechanism for learning representations.
Standard ConvNet architectures are usually composed of a set of feed-forward 
operations, with convolutional layers 
followed by fully connected layers.
The first convolutional layers capture some basic features like color, gradient strength, edge orientation, etc., while the fully connected layers extract more abstract, 
complex semantic features \cite{Razavian14}. In addition, \cite{Razavian14} 
indicates that ConvNet extracted features outperform hand-crafted ones.       
      
We investigate three methods to geolocate our images automatically. To this end we evaluate three ConvNets variants~\cite{Chatfield14, Krizhevsky12} as described below. We focus on the case when the available data is not sufficient to train a deep ConvNet from scratch.

\subsubsection{Pre-trained ConvNet + SVM (Pretrained+SVM)}
\label{sec:method1}

This method is inspired by \cite{DeCAF_Donahue}. The main idea is that given 
an input image $I$ and a pre-trained ConvNet $f$, when $I$ is pushed through
$f$, every layer of $f$ produces an activation response. 
The activations at a specific layer(s) are regarded as the features $x$ of an 
input image, $x = f(I)$.
Then, having a set of image - label pairs $(I_i,y_i)$, we extract the 
features $x_i$ from each image. Using the feature - label pairs $(x_i,y_i)$, 
we train a multiclass classifier $g$, i.e. a Multiclass Support Vector Machine 
(SVM), used at test time to predict the class label $\hat{y_i}$.

Following this methodology, the pre-trained network $f$ becomes a feature 
extraction mechanism, on top of which a SVM classifier $g$ is used to assign
city labels to human-centered images. 
Hence, $\hat{y_i} = g( x_i ) = g(~ f( I_i ) ~) $.

%


\begin{figure}
\includegraphics[width=0.48\textwidth]{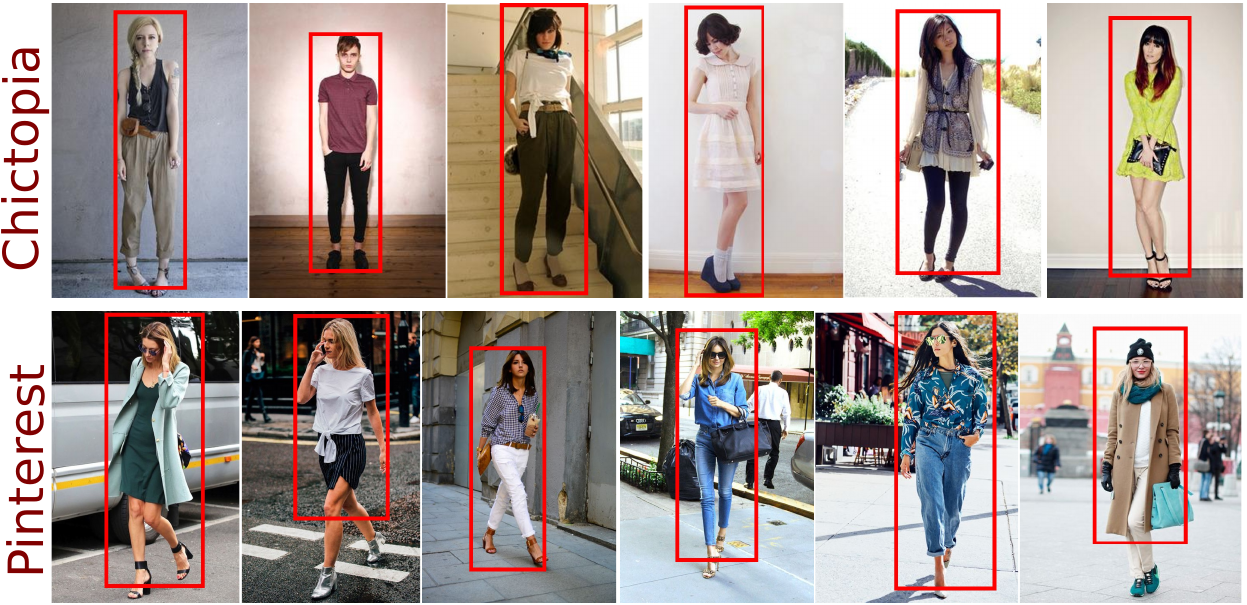}

\caption{Example images from the Fashion 144k/Chictopia (top) and Pinterest (bottom) datasets. We show in red the bounding boxes produced by the Faster R-CNN detector~\cite{Ren15} used in the human-based feature pooling experiment.}
\label{fig:datasetExamples}
\end{figure}

\subsubsection{Fine-tuning a pre-trained ConvNet (Finetuned)}
\label{sec:method2}

A deep neural network has millions of parameters to tune, 
which means that it will need a huge dataset in order to set
these parameters properly. For instance, for ``VGG-F''~\cite{Chatfield14} 
138GB of image data were used for training, 6.3GB of image data 
for validation and 13GB of image data for testing. 

Fine-tuning is an alternative for situations when a large 
dataset is not available to train a network from
scratch. Moreover, it has already been proved to 
yield a better performance than when training a network  
from scratch with insufficient data \cite{Lin15}. 
It follows the same architecture as the pre-trained model $f$, but changes the last layers to satisfy the new classification task, thus producing 
a new network $f'$. Different from the previous method, given 
an image $I_i$, we focus on the output of $f'$~(over 
the set of classes of interest) as the predicted class label
$\hat{y_i}=f'(I_i)$.


\subsubsection{Fine-tuned ConvNet + SVM (Finetuned+SVM)} 
\label{sec:method3}

This method is a combination of the previous two methods.
It follows a similar procedure as \textit{Pretrained+SVM}, i.e. 
the class labels are predicted as $\hat{y_i} = g( x_i ) = g(~ f( I_i ) ~) $.
However, different from \textit{Pretrained+SVM}, here we replace 
$f$ by its fine-tunned counterpart $f'$~(\textit{Finetuned})
and use activation responses from $f'$ as features.
Hence, $\hat{y_i} = g( x_i ) = g(~ f'( I_i ) ~) $.

\subsection{Inspecting Features Learned by the Network}
\label{sec:relevantFeatIdentification}

One of the main strengths of deep models is their ability to learn features that 
produce high performance for a task of interest. This is achieved by iteratively 
modeling abstract concepts through ensembles of simpler ones.
Looking at these simpler concepts, i.e. internal activations, can provide an 
insight on the visual features that the deep model is taking into account when 
making predictions. 
Further processing of these internal activations will allow us to reach some 
understanding of what the model is actually looking at when classifying our 
human-centered images.
To this end, we analyze the proportion of activations within the image region depicting the persons.
The objective is to identify whether the features considered during 
the prediction come from either the foreground, i.e. the person 
depicted in the image, or the background, i.e. the context.
Towards this objective, we compute the proportion between 
the feature response within the bounding box around the person with 
respect to the response on the whole image. 
Then we estimate the probability density function per filter.
We expect that features correlated with the persons will have 
higher proportion than those in the context.

As discussed in Section~\ref{sec:relatedWork}, there are two families
of methods that can be followed in order to visualize the strength
of internal activations. 
On the one hand, methods based on DeconvNets~\cite{gruen16,Simonyan13,Springenberg14,Zeiler14} generate 
heatmaps by backpropagating the activations on given layer-filter 
location back to the input image space. These methods produce detailed 
pixel-level visualizations at the cost of additional computations.
On the other hand, methods based on upsampled activation maps~\cite{netdissect2017,zhou2015cnnlocalization} generate 
relatively coarser visualizations with the advantange of only 
requiring a relatively simpler additional upsampling operation.
Based on these observations, we follow the line of work from ~\cite{netdissect2017,zhou2015cnnlocalization} and generate ``heatmaps'' 
of internal features by upsampling the activations of every 
filter in the last convolutional layer. More specifically, we focus 
on the activation maps produced after the Rectifier Linear Unit (ReLU) 
operation.
Then, from these heatmaps we compute the proportion of activations 
within the region around the person depicted in the image.

\section{Evaluation}
\label{sec:evaluation}

In the following section we present the protocol followed to
investigate the problem at hand.
Then, we present two directions to address this problem, i.e. 
a user study (Section~\ref{sec:userStudy}) and a series of 
experiments based on automatic methods (Sections~\ref{sec:imageFeaPooling}, 
\ref{sec:humanFeaPooling}, \ref{sec:classificationRetrievalComparison} \& 
\ref{sec:largeScaleExp}). We conclude this section with a 
deeper inspection of the features learned by the networks~(Section~\ref{sec:relevantFeatInspection}).

\textbf{Datasets:}
We use three different datasets in our experiments. 
The first one is the Fashion 144k dataset~\cite{SimoSerra15} which contains 144,169 images 
posted from the largest fashion website \textit{chictopia.com}. These images
are "human-centered": a photograph of the person who posted it wearing an outfit, 
with an outdoor or indoor background. There are 3,443 different photographing 
locations but the number of photographs of each location varies considerably: some 
locations like Los Angeles have thousands of posts while most locations only 
contain less than 100 images. To keep the data balanced, we chose 
12 different locations with more than 1,000 images. This produced 
a dataset composed by 12,448 images covering 12 city classes, i.e. 
'LA', 'London', 'Madrid', 'Melbourne', 'Miami', 'Montreal', 'Moscow', 
'North Europe region', 'NYC', 'Paris', 'San Francisco' and 'Vancouver'.
The second and third datasets are quite similar. We collected them 
ourselves from \textit{chictopia.com} and \textit{pinterest.com}. 
Both contain the same 12 locations, with 13,332 and 12,671 
images in total, respectively. In terms of photographing style, Pinterest images represent "instant" photos taken on the fly, while images from Chictopia are more planned, i.e. a user is posing (like a model) for the photo.
See Figure~\ref{fig:datasetExamples} for some example images
from these datasets.
In addition, we define a \textit{Mix} dataset that 
combines our two self-collected datasets (i.e. based on Chictopia \'and Pinterest).

We divide each dataset into three independent parts: 
70\% for training, 15\% for validation and 15\% for testing.


\textbf{Performance metric:}
After training the classifiers, we evaluate their performance on the testing dataset. 
The mean accuracy for the 12 locations was calculated and we use this mean class accuracy (mCA) 
as the performance metric.

\subsection{Exp.1: Human Performance Study}
\label{sec:userStudy}


We start our evaluation by performing a study on 
how well people perform the task of geolocation 
from human-centered images. It will also provide 
an indication for the difficulty of the geolocation 
task.

\begin{figure}
\centering
\includegraphics[width=0.46\textwidth]{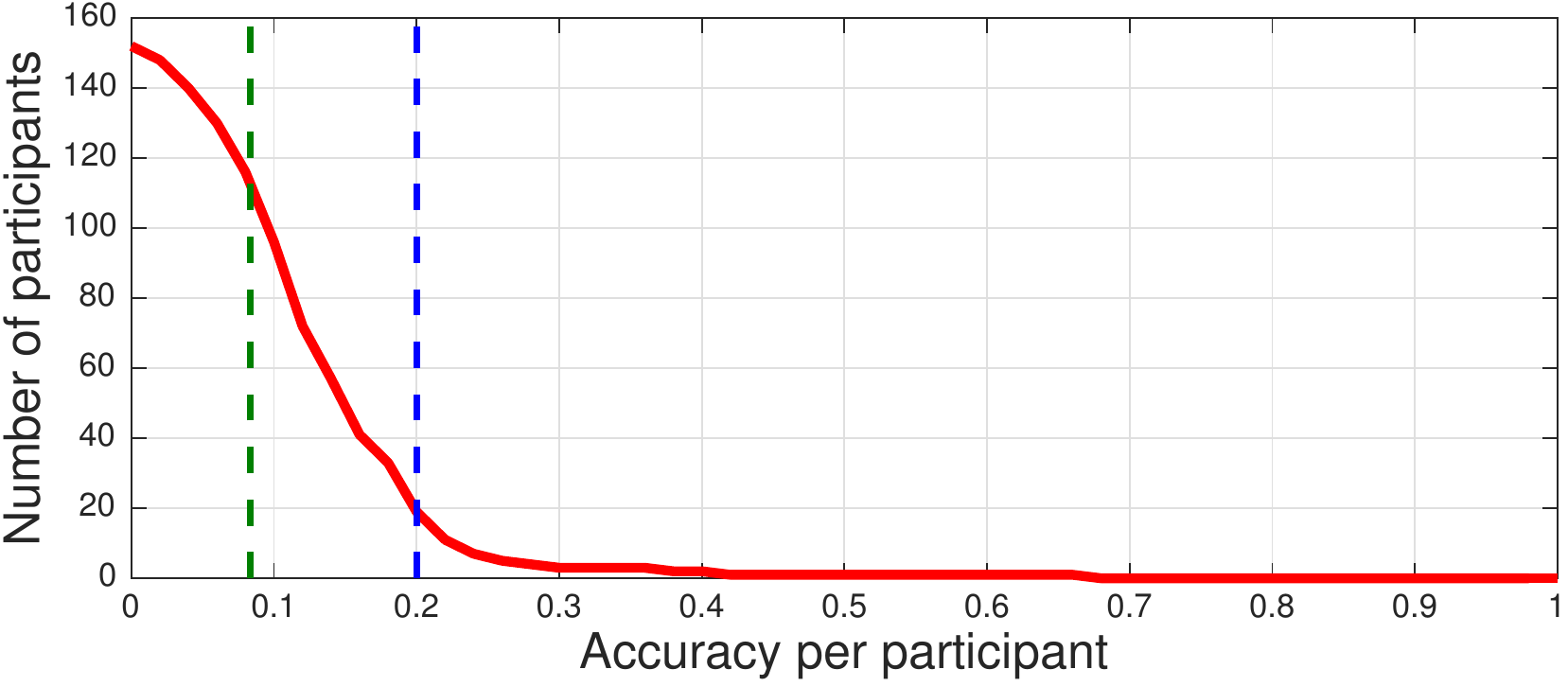}
\caption{Cumulated score distribution over all participants (red). Chance-level performance (green).
Note that more than $87\%$ of the participants obtained an accuracy below 0.2 (blue).
}
\label{fig:participantScoreDistribution}
\end{figure}

We conducted a survey asking 
people to determine where a given photo was taken. 
We randomly select 24 images from each city from our 
self-collected datasets~(Chictopia and Pinterest), 
producing a total of 288 images for our online 
questionnaire. 
Each time, one image is presented to the participant 
and the participant is asked to select one city from 
the list of 12 possibilities.    
In total, we received 6,505 responses from 153 participants
with ages between 20 - 30, with an average of $\sim$23 votes 
given for every image.
Among these responses, 3,258 are for the images from the 
Chictopia dataset while 3,247 are for the Pinterest dataset. 
\jose{
In addition, we conducted an \textit{extended} survey in 
which four participants were given access to annotated 
training images.
}
We calculated the mean class accuracy (mCA) from those responses. 
The quantitative results can be found in Table~\ref{tab:us}. 
For reference, we present in Figure \ref{fig:participantScoreDistribution}, 
the cumulative distribution of the accuracy obtained by the 
participants of the survey.

\begin{table}
\centering
  \caption{Quantitative results for user study. Mean class accuracy (mCA) on the Chictopia and Pinterest based datasets.}
  \label{tab:us}
  \begin{tabular}{lccc}
    \toprule
    & Chictopia&Pinterest & Mix \\
    \midrule
    Human  & 11.60 & 12.29 & 11.94 \\
    Human (extended) & 23.86 & 18.75 & 21.20 \\
  
  \bottomrule
\end{tabular}
\end{table}

\textbf{Discussion:}
\jose{
Results from our survey suggest that performance between 
female and male participants is similar (0.39\% difference) on 
the geolocation task.
}
From Table~\ref{tab:us}, we can see that for the initial survey,
the human performance on Pinterest dataset is slightly
higher than on Chictopia. The reason might be because 
the photos of Pinterest tend to offer a bit more background 
context. Users may gain more cues from the 
background to determine the geographic location. 
It is remarkable that, for this survey, on 
both datasets human performance is only slightly 
better than a random guess over 12 classes (8.33\%). 
In addition, we notice that for the case of the \textit{extended}
survey, human performance increases to $\sim$0.21 mCA.
This suggests that the task is difficult. This is further 
stressed by Figure~\ref{fig:participantScoreDistribution} (blue line), 
where it is noticeable that more than $87\%$ of the participants 
obtained an accuracy below 0.2. Moreover, we can notice 
a very small number of participants reaching accuracy 
scores above 0.5.

\subsection{Exp.2: Automatic Geolocation via ConvNets}
\label{sec:imageFeaPooling}

%
%

In this section we evaluate the performance of the automatic
methods presented in Section~\ref{sec:imageClassification}.
We select "VGG-F"~\cite{Chatfield14} as  architecture 
for the implementation of our automatic methods. VGG-F is a 
feed-forward 21-layer ConvNet with 15 convolutional layers,
five fully connected layers and one softmax layer. 
For the case of \textit{Pretrained+SVM}~(Section~\ref{sec:method1}), 
considering the activation from internal layers as a feature, 
we take the activations from the last fully connected layer~(\textit{fc7})
of a VGG-F network pretrained on ImageNet.
This produces a feature vector with length 4,096 per image that 
we use to train a multiclass SVM. 
Cross-validation is adopted to get the best parameters for the SVM.

%
For the case of \textit{Finetuned}~(Section~\ref{sec:method2}),  
the dimension of the output layer is modified to 12 instead 
of the original 1,000 in order to produce an output focused on our 
classes of interest. The weights of this last layer are 
initialized with random values from a Gaussian distribution. 
Additionally, two more dropout layers are added between 
\textit{fc6} and \textit{fc7} and between \textit{fc7} and \textit{fc8}. 
We tested two fixed learning rates (1e-4 and 1e-5) 
and a range of adaptive rates (1e-8 to 1e-4) to find the best one.
For the case of \textit{Finetuned+SVM}~(Section~\ref{sec:method3}), 
we first fine-tune the network, as done for \textit{Finetuned}, 
and use the fine-tuned network as feature extraction mechanism~(\textit{Finetuned+SVM}).

During training, for the case of our self-collected Chictopia 
and Pinterest datasets, we use the \textit{Mix} dataset to train 
ConvNets/classifiers by adopting the above three methodologies 
and evaluate the ConvNets/classifiers on each of the subsets, i.e. 
Chictopia, Pinterest and the \textit{Mix} dataset, respectively. 
For the case of the Fashion 144k dataset~\cite{SimoSerra15}, we use the 
pre-defined image sets for training and testing.
All our models\footnote{Publicly available at \url{http://github.com/shadowwkl/An-Analysis-of-Human-centered-Geolocation}} are trained using the MatConvNet 
framework~\cite{MatConvNet}.
Table~\ref{tab:imageBased} shows the quantitative performances for this 
experiment.



\begin{table}
  \caption{Mean class accuracy~(mCA) of image-based feature pooling in percentage points}
  \label{tab:imageBased}
  
  \resizebox{\columnwidth}{!}{%
  \begin{tabular}{lccc|c}
    \toprule
    & Chictopia&Pinterest & Mix & Fashion 144k \\
    \midrule
    Pretrained+SVM & 33.97  & 25.22  & 29.94   & 37.16 \\
    Finetuned & \textbf{40.75}  & \textbf{28.13}  & \textbf{34.75}   & \textbf{39.15} \\
    Finetuned+SVM & 35.45  & 24.25  & 30.26   & 35.60 \\
    Human & 11.60 & 12.29 & 11.94 & --\\
    Human (extended) & 23.86 & 18.75 & 20.92 & -- \\
  \bottomrule
\end{tabular}
}

\end{table}

\textbf{Discussion:}
From Table~\ref{tab:imageBased}, we can observe that 
the automatic methods perform the best on the Chictopia dataset. 
Moreover, their performance is substantially higher than human 
performance, and this seems to be a trend for all three datasets.
On the Chictopia dataset, the automatic methods have a much 
higher mCA than human whose mCA is $\sim$21 percentage 
points (pp), for the \textit{extended} survey. 
\textit{Finetuned} has the best performance 
among the three automatic methods with its highest mCA 40.75\%. 
\textit{Finetuned+SVM} follows with 35.45\% mCA while \textit{Pretrained+SVM} 
has the lowest mCA (33.97\%). 
Therefore, the features extracted from the fine-tuned model 
yield a better performance than the pre-trained model when 
using the same classification technique.
Overall these results show that predicting the geographic 
location where the analyzed "human-centered" photos are taken 
is - to some extent - possible.

\subsection{Exp.3: Human-based Feature Pooling}
\label{sec:humanFeaPooling}



In the previous experiment we followed an image-based 
feature pooling approach, i.e. features were extracted by 
considering the whole image as input.
Since the context information from the background may serve as a
strong cue for determining the geographic location, in this 
experiment we investigate to what extent the problem can 
be solved if we only pool features from the persons appearing 
in the images. 
Towards this objective, we apply the Faster R-CNN~\cite{Ren15} 
detector to localize the person appearing in the picture (See Figure~\ref{fig:datasetExamples}). 
After the detection, we clip the image according to the predicted 
bounding box and obtain image regions focusing on the individuals 
appearing on the images. These image regions will now serve as 
input to the network. 
We perform this process on every image from the Chictopia and Pinterest 
datasets. We refer to this procedure as human-based feature pooling.

Similar to Section~\ref{sec:imageFeaPooling}, we evaluate the same three 
methods with the difference that we feed the image regions produced by the 
Faster R-CNN as inputs.
The quantitative results of this experiment are shown in Table~\ref{tab:humanBased}.


\begin{table}
  \caption{Mean class accuracy~(mCA) of human-based feature pooling in percentage points}
  \label{tab:humanBased}
  
  \resizebox{\columnwidth}{!}{%
  \begin{tabular}{lccc|c}
    \toprule
    & Chictopia & Pinterest & Mix & Fashion 144k \\
    \midrule
    Pretrained+SVM & 28.07 & 19.54 & 24.06 & 33.28 \\
    Finetuned & \textbf{35.00} & \textbf{22.00} & \textbf{28.79} & \textbf{35.20} \\
    Finetuned+SVM & 29.66 & 19.26 & 24.69 & 30.44 \\
  
  \bottomrule
\end{tabular}
}

\end{table}



\begin{figure}
\centering
\includegraphics[width=0.5\textwidth]{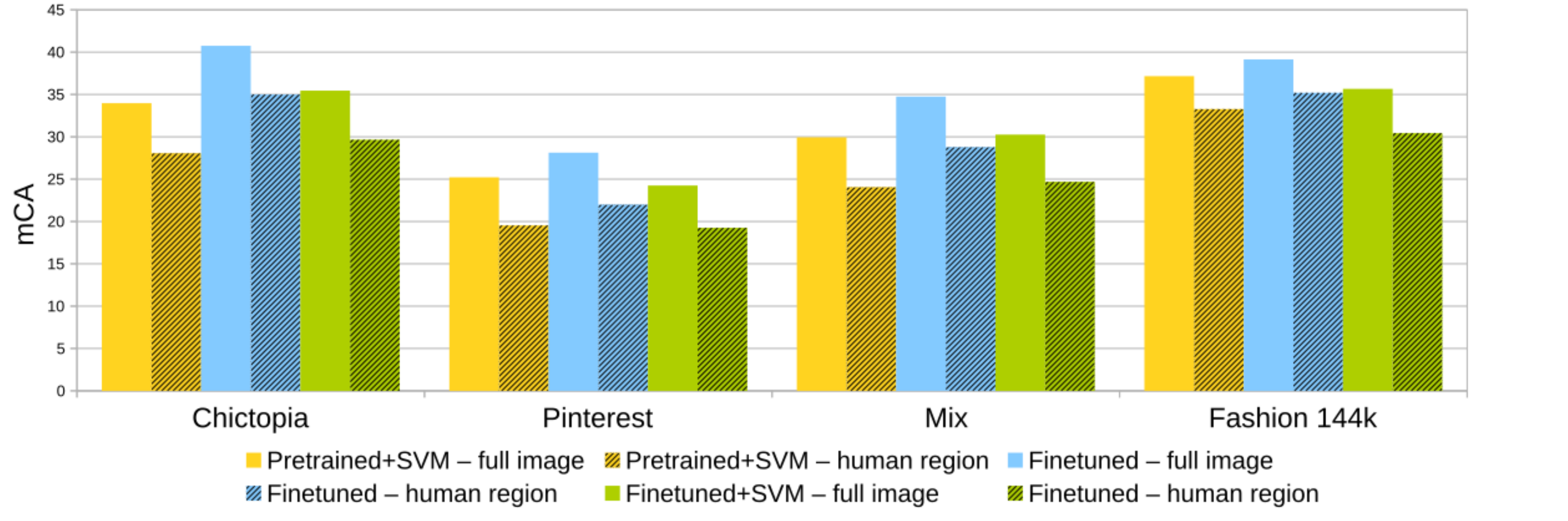}
\caption{Quantitative Performance Comparison.}
\label{fig:quantitativeComparison}
\vspace{-4mm}
\end{figure}


\textbf{Discussion:}
From Table~\ref{tab:humanBased}, we again see that \textit{Finetuned} 
performs best on the Fashion 144k (35.20\%) and Chictopia (35\%) datasets, 
but also that the accuracy drops about four percent compared to Exp.2. 
By comparing the quantitative performance between human-based feature pooling and image-based feature pooling (Section~\ref{sec:imageFeaPooling}), we can see a general trend of performance decrease by four to six percent (Figure~\ref{fig:quantitativeComparison}). 
This shows that the context information does play a role in determining the final 
decision, but that it is not critical since the performance after removing 
most background information is still two times better than that of humans.




\subsection{Exp.4: Classification vs. Retrieval -based Human-centered Geolocation}
\label{sec:classificationRetrievalComparison}

\begin{figure}[t]
\centering
\includegraphics[width=0.5\textwidth]{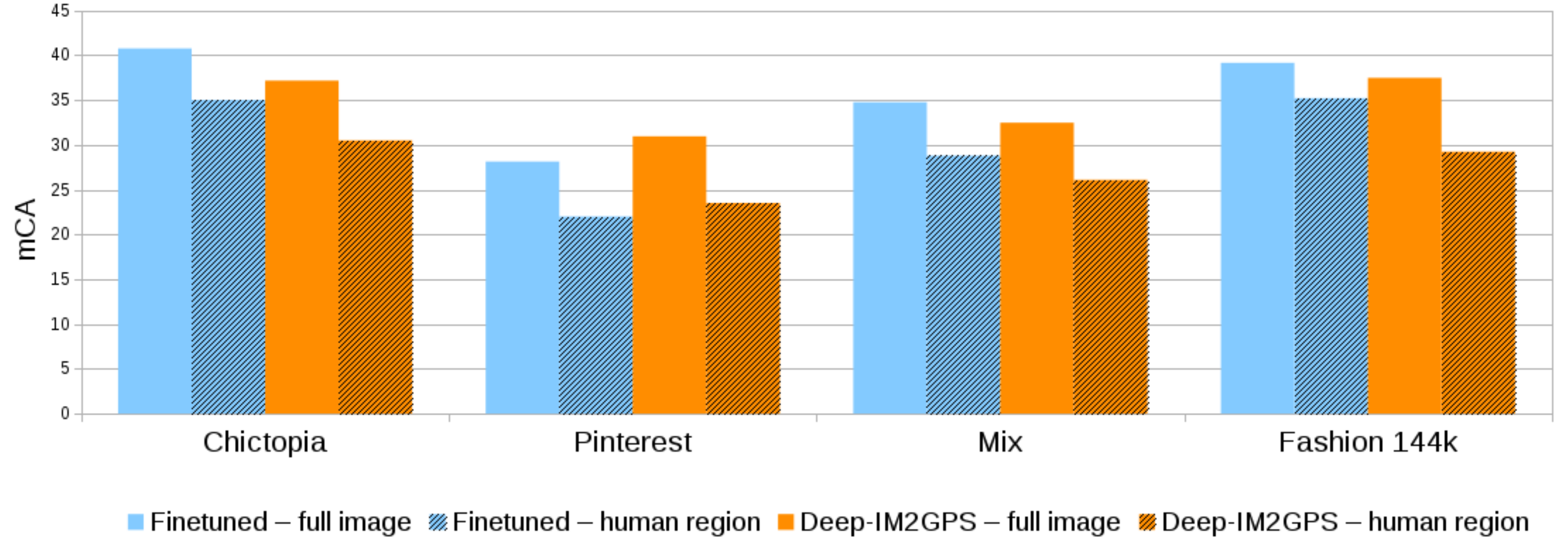}

\caption{Classification vs retrieval -based geolocation.}
\label{fig:classificationVSretrieval}
\end{figure}

In its current form our \textit{Finetuned} variant could be considered a simplified version of the classification-based method from \cite{Weyand16} where instead of predicting discrete GPS coordinates our model predicts city classes. 
In order to provide some insight on the performance that retrieval-based methods can achieve on the geolocation of human-centered images,  we evaluate an additional method following the retrieval-oriented IM2GPS approach proposed recently in \cite{VodeepIM2GPS}. In this case, all the examples from the training set are encoded using the \textit{Finetuned} network and considered as reference data for k-Nearest Neighbors (k-NN) search within IM2GPS. In addition, since we do not predict GPS-related labels as output, we modify the output to produce a weighted vote of the class (city) labels corresponding to the retrieved k-NN examples from the reference set. We refer to this method as \textit{Deep-IM2GPS}.
We compare the performance of this \textit{Deep-IM2GPS} variant w.r.t. our top performing classification-based method \textit{Finetuned}. We report quantitative results in Figure~\ref{fig:classificationVSretrieval}.


\textbf{Discussion:}
A quick inspection to Figure~\ref{fig:classificationVSretrieval} reveals that  classification-based methods are almost always better than their retrieval-based counterparts except for the case of the Pinterest dataset. 
We hypothesize that this might be related to the fact that images from the Pinterest dataset have more contextual information and that retrieval-based methods might be better suited to exploit this type of feature.\\
%
In addition, we notice that also for \textit{Deep-IM2GPS} it is still clear 
the trend that classification performance is superior when using image-based 
feature pooling~(Section~\ref{sec:imageFeaPooling}).


\subsection{Exp.5: Considering a larger number of classes }
\label{sec:largeScaleExp}

\begin{figure}
\centering
\includegraphics[width=0.48\textwidth]{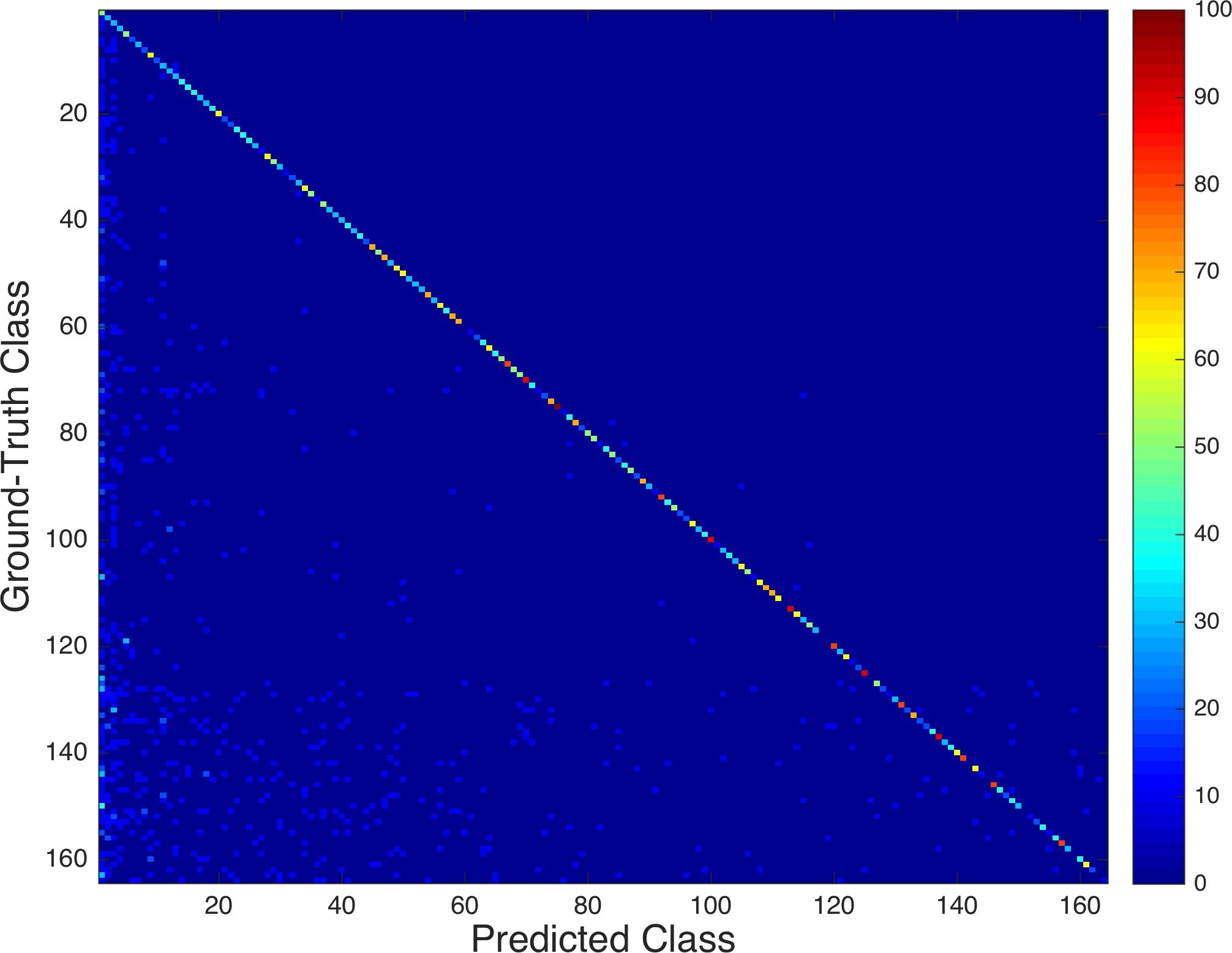}
\caption{Confusion matrix from the \textit{Finetuned} model considering 164 locations (classes).}
\label{fig:largeScaleExpConfMat}
\end{figure}

In order to verify whether our observations hold when considering a larger number of classes, we consider the cities from the Fashion144K dataset~\cite{SimoSerra15} that have more than 100 images examples per class.
This leads us to a set of 70k images covering 164 cities (classes).\\
Based on the results obtained in the previous experiment, we take the top-performing automatic method, i.e. Image-based Feature Pooling with a \textit{Finetuned} Network architecture.
In addition, we report performance for the retrieval-based method \textit{Deep-IM2GPS}~\cite{VodeepIM2GPS}.\\
We present quantitative results in Table~\ref{tab:largeScaleExp}. 
For completeness, we report performance on both geolocation methods, i.e. \textit{Finetuned} and \textit{Deep-IM2GPS}, when using image-based feature pooling (\textit{Image-bFP}) and 
human-based feature pooling (\textit{Human-bFP}).
The confusion matrix of our \textit{Finetuned} method is presented in Figure~\ref{fig:largeScaleExpConfMat}. 
\jose{
Classes are sorted by the number of examples they contain in decreasing order.
}

\textbf{Discussion:} For the case of the \textit{Finetuned} method combined with image-based feature pooling, we notice a drop of ~1\% when compared with the experiment that considers 12 cities~(Table~\ref{tab:imageBased}). 
\jose{
This low difference is somewhat surprising. A deeper inspection of the data revealed that as we go to classes with lower numbers of examples, i.e. more remote locations, the number of users uploading images from those locations is significantly smaller. Thus, even when additional classes are added, classifying these is a simpler problem.
We can verify this in Figure~\ref{fig:largeScaleExpConfMat} where classes with lower number of examples (larger class-id) seem to have superior performance.
}
We notice that in this larger-scale experiment the classification-based method, \textit{Finetuned}, still outperforms its retrieval-based counterpart, \textit{Deep-IM2GPS}~\cite{VodeepIM2GPS}. 
In addition, in this significantly more complex large-scale experiment the \textit{Human} performance initially obtained for the 12-class experiment (Section~\ref{sec:userStudy}) is expected to have a significant drop.
Finally, the trend that image-based feature pooling (\textit{Image-bFP}) provides superior performance over its human-based counterpart (\textit{Human-bFP}) is still clear.


\begin{table}
\centering
  \caption{Mean class accuracy~(mCA) in percentage points for the large-scale experiment (164 classes) on the Fashion144k dataset.}
  \label{tab:largeScaleExp}
  
  \resizebox{0.8\columnwidth}{!}{%
  \begin{tabular}{lccc}
    \toprule
    & Image-bFP & Human-bFP \\
    \midrule
    Finetuned & \textbf{38.26} & \textbf{35.16} \\
    Deep-IM2GPS~\cite{VodeepIM2GPS} & 33.49 & 24.88  \\
  
  \bottomrule
\end{tabular}
}

\end{table}


\begin{figure*}
\centering
\includegraphics[width=0.86\textwidth]{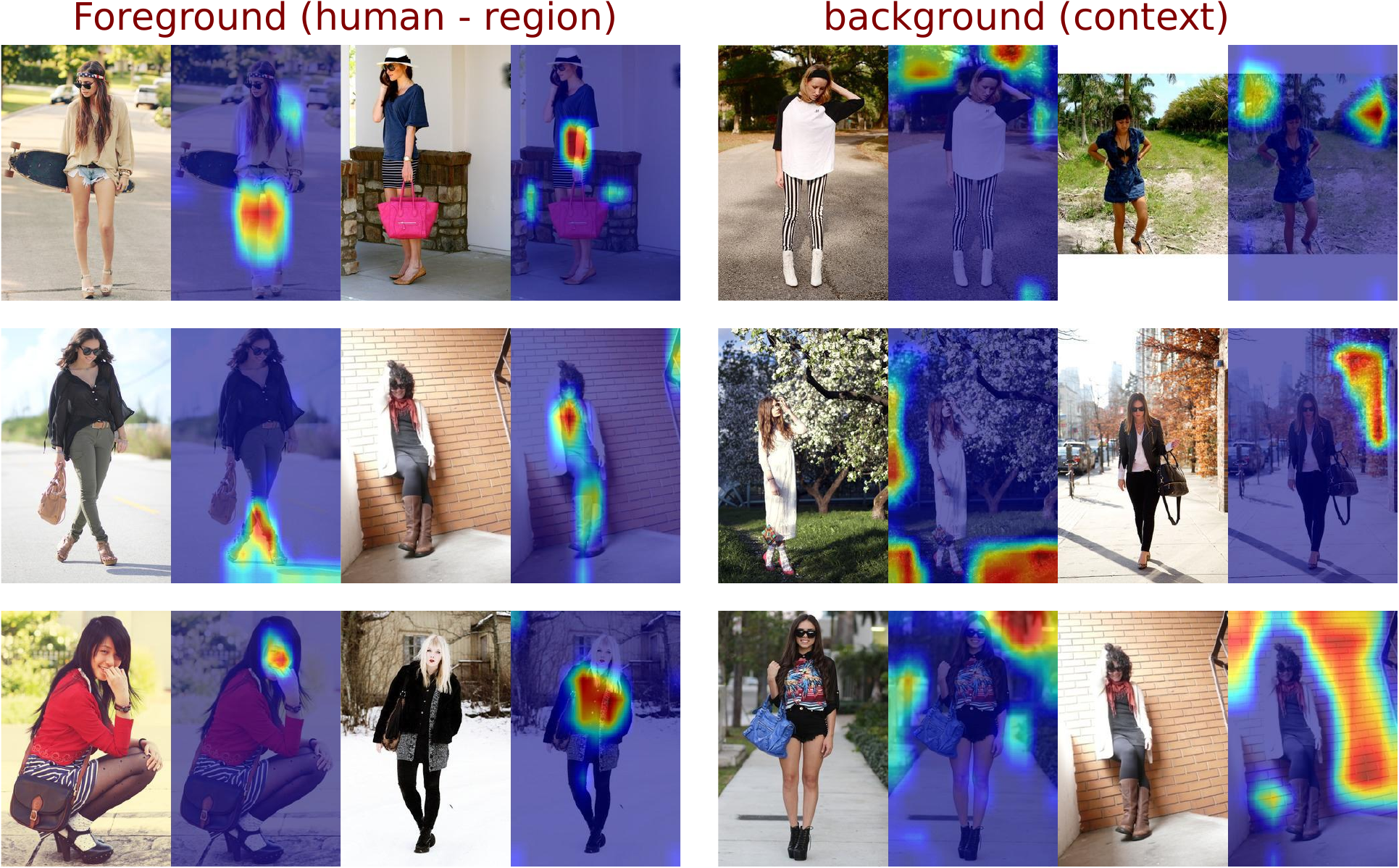}

\caption{Qualitative examples from upscaled activation maps from the last convolutional layer of our \textit{Finetuned} model. Visualizations are grouped based on the origin, i.e. foreground (human-region) or background (context), of the visual cues they model.}
\label{fig:relevantFeatures}
\end{figure*}

\subsection{Exp.6: Verifying the Origin of the Features Learned by the Network}
\label{sec:relevantFeatInspection}




In this experiment we further investigate the origin, 
i.e. foreground~(human-region) or background~(context),
of the visual cues considered by the network. 
We take a deeper look at the visual patterns that the 
automatic models take into account when predicting geographic locations. 
More specifically, we focus on the network activations in the last 
convolutional layer from \textit{Finetuned} trained on the Fashion 144k dataset based on image-based feature pooling (Section~\ref{sec:imageFeaPooling}).
In addition, following the procedure described in 
Section~\ref{sec:relevantFeatIdentification}, we analyze 
the proportion of activations of the features within the 
bounding boxes predicted by the Faster R-CNN detector~\cite{Ren15}.

\begin{figure}
\centering
\includegraphics[width=0.44\textwidth]{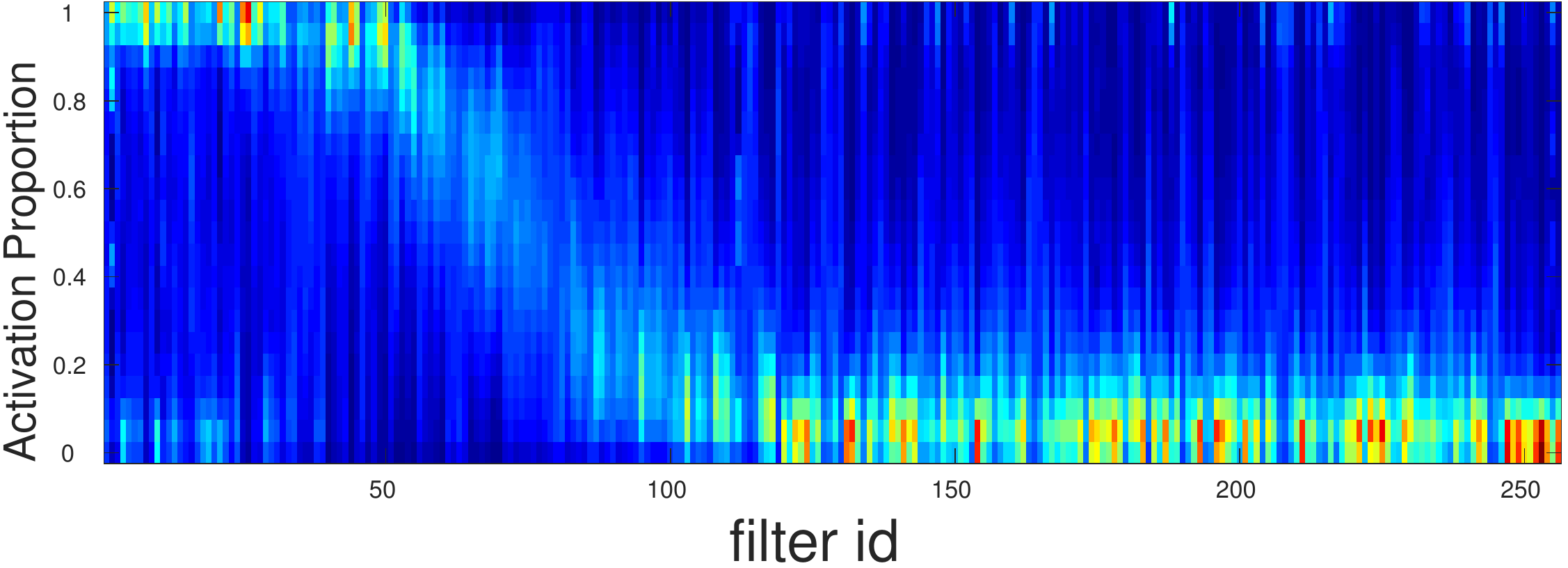}
\caption{Density of the proportion of the activations of the features (from the last convolutional layer) within the image region where persons appear. Filters are sorted by their activation proportion, from high to low, along the x-axis.}
\label{fig:filterBBcoverage}
\vspace{-6mm}
\end{figure}

In Figure~\ref{fig:filterBBcoverage}, we show the 
probability density of this proportion color coded in 
jet scale. For clarity, the features are sorted in the 
order of decreasing proportion from left to right along the x-axis.
In the y-axis we indicate the proportion of activations 
occurring within bounding box.
Finally, Figure~\ref{fig:relevantFeatures} shows 
qualitative visualizations of the features considered 
by the last convolutional layer of the network grouped
by their origin.

\textbf{Discussion:}
%
%
Going back to the question of whether these relevant 
features lie on the background (context) or on the foreground 
(human-region), Figure~\ref{fig:filterBBcoverage} shows
that these features are found grouped in two clusters. 
On the top-left corner, we can note a group of activations for 
filters that occur mostly in the foreground. We can notice group 
of features (filter id: 1-60) with more than 80\% 
occurrence on the foreground.
In the bottom-mid-right, we can see a spread-out group (filter id: 120-256). 
These are activations of filters that lie in the background. 
Finally, on the center (filter id: 60-100), we can notice 
a set of activations from filters that are almost equally 
shared between foreground and background.
It is visible that these shared features have a lower intensity
that those located either on the foreground (top-left) or 
in the background (bottom-right).

To complement the observations made previously, we qualitatively
evaluate the features considered by the network in the last 
convolutional layer. In Figure~\ref{fig:relevantFeatures} we 
visualize some of the upscaled activation maps from these features. 
We can notice that some features seem to be directly related 
to the individuals present in the image (Figure~\ref{fig:relevantFeatures}~(left)). For instance, for ’L.A.’ and ’Miami’, we see filters with strong activations near faces, arms, and uncovered knees/legs.
This implies the fact that these two locations are relatively warmer. 
Likewise, we can notice that some other filters have strong 
activations related to clothing, some fire on jackets, scarfs, 
and nearby shoes. For instance, in ’Paris’ and 'Moscow' beige 
and blue colors seem to be popular for clothing, respectively.
In Figure~\ref{fig:relevantFeatures}~(right), we can clearly see 
that some of the learned features focus on aspects from the context.
For instance, for ’Melbourne’ and 'North Europe' green and white color features highlight vegetation and snow that appear in their respective landscapes. 
It is likely that there is a correlation between these 
features and the classes (cities) modelled by the network. 
This shows that indeed, there are some features on human-centered 
images that can be informative for geographic localization.
In future work, we will take a deeper look and analyze
where these relevant features come from, either from the 
physical characteristics of the persons 
or from the clothing they wear. 
In this regard, methods of clothing parsing/segmentation \cite{Liu15,Yamaguchi12}
might be a possible direction to achieve "clothing-based" 
feature pooling.
\jose{
Moreover, we will further investigate whether the scene 
type, i.e. indoors vs. outdoors, has an effect on automatic 
visual geolocation performance.
}

\section{Discussion}
\label{sec:discussion}

In this paper we analyze the problem of geolocation from images by looking at the people therein. Our main objective is to analyze the task from this particular aspect of image content, but it stands to reason to do the same for e.g. buildings, vegetation, text, etc. that may be present, and then exploit whatever useful class that is available. We believe that such class-specific avenue will lead to improved results for a problem that is considered important (e.g. the subject of a DARPA research program). Given that this type of human-centered data has been rarely considered for geolocation - although it is quite instructive according to our results - our analysis at the same time produces a baseline to foster such work. The contribution of the paper does not lie in its technical novelty, and no claim to that effect was made. Further possible applications of the evaluated methods that were mentioned, will be examined as part of future work. 

In Section~\ref{sec:userStudy}, we conducted a Web survey and let people determine the location of a given image. 
Partcipants of the survey obtained a quite low mCA of $\sim$12\%~ (Table~\ref{tab:us}). This reflects the difficulty of the problem when focusing on human-centered images, a good portion of which are indoors and/or provide arbitrary background information.
The large disparity between human performance and the automatic methods can be attributed to the lack of training or expertise that the participants of the survey have when compared to the automatic methods. While the participants of the survey were asked to provide a guess on the location of an image, and in some cases had access to training images, the automatic methods had the advantage of observing several thousands of training images in advance.
Moreover, when inspecting the confusion matrix computed from the results of the Web survey, we noticed that there were some cities that were usually confused, e.g. ('Miami', 'LA');  ('Paris', 'London'); ('North Europe', 'Moscow').  
This might be caused by a perceived similarity between these cities.

Perhaps, with a longer "training" time for the surveyed participants human performance might be increased. However, given some of the similarities between some classes and the lack of popular landmark information,  we expect this improvement to be insufficient to reach that of the automatic methods. A similar observation was reached in the user study conducted in \cite{Weyand16}.
Moreover, as observed in the user study conducted in \cite{Mehta16}, when performing visual geolocation, humans mostly rely on natural cues (e.g. sun position, animal types, natural landmarks, etc.) as well as man-made structures (e.g. architecture, road signs, traffic rules) which are significantly reduced in the human-centered images that are the focus of our analysis. It was observed that some mistakes were caused by erroneous preconceptions used by some participants. For instance, some participants of our Web survey mentioned that they were looking for a beach scene for the 'Miami' class. Likewise, there was an expectation of finding snow in classes related to Canada, i.e. 'Montreal' and 'Vancouver'. These initial preconceptions could bias the decision of the participants and affect the overall human performance given the fact that the occurrence of the mentioned cues are almost non-existent on the images of the mentioned classes.
Please refer to \cite{Mehta16} for a detailed description on human factors for visual geolocation.

\section{Conclusion}
\label{sec:conclusion}

We have investigated the problem of predicting the geographic location 
where a human-centered photo was taken.
We have conducted an analysis of several 
aspects to this challenge. 
Our results suggest that it can be resolved successfully 
to some extent in an automatic fashion, which even surpasses human 
performance.
A close inspection of the trained models shows that indeed, there 
are some human-centered characteristics, e.g. clothing style, physical 
features, accessories, which are informative for the task.
Moreover, it reveals that, despite their apparent low occurrence, 
contextual features, e.g. wall type, natural environment, etc., 
are also taken into account by the automatic methods.

\vspace{2mm}
\small{
\noindent\textbf{Acknowledgments:}
This work was partially supported by the KU Leuven PDM Grant PDM/16/131, the KU Leuven GOA project CAMETRON,
and a NVIDIA Academic Hardware Grant.
}

{\small
\bibliographystyle{ieee}
\bibliography{egbib}
}

\end{document}